%% file: main.tex
\documentclass[10pt,twocolumn,letterpaper]{article}

\usepackage[pagenumbers]{iccv} 

\input{preamble}

\definecolor{iccvblue}{rgb}{0.21,0.49,0.74}
\usepackage[pagebackref,breaklinks,colorlinks,allcolors=iccvblue]{hyperref}

\def\paperID{***} 
\def\confName{ICCV}
\def\confYear{2025}

\title{\textit{What Changed?} Detecting and Evaluating Instruction-Guided Image Edits\\with Multimodal Large Language Models}

\author{Lorenzo Baraldi$^{*2}$, Davide Bucciarelli$^{*1,2}$, Federico Betti$^{*3}$,\\ Marcella Cornia$^{1}$, Lorenzo Baraldi$^{1}$, Nicu Sebe$^{3}$, Rita Cucchiara$^{1,4}$\\
$^1$University of Modena and Reggio Emilia, Italy \quad$^2$University of Pisa, Italy\\
$^3$University of Trento, Italy \quad$^4$IIT-CNR, Italy \\
{\tt\small $^1$\{name.surname\}@unimore.it, $^2$\{name.surname\}@phd.unipi.it, $^3$\{name.surname\}@unitn.it}
}

\begin{document}

\begin{figure}[htb]
\twocolumn[{
\renewcommand\twocolumn[1][]{#1}%
\maketitle
\vspace{-25pt}
\includegraphics[width=0.99\linewidth]{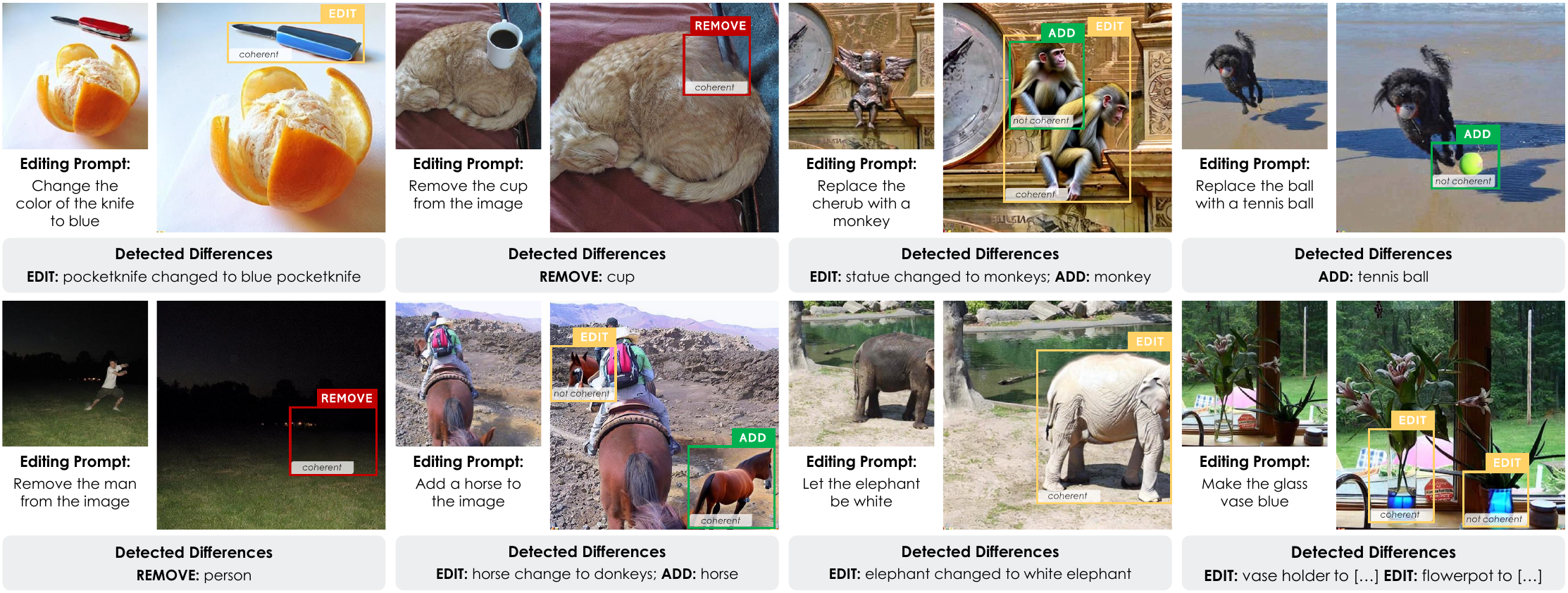}
\vspace{-0.2cm}
\caption{Qualitative examples from \ours. Our approach detects differences between an original image and an edited one, identifying the involved objects and the type of edit. Further, \ours evaluates each difference to determine its coherence with the editing prompt.}
\label{fig:first_page}
\vspace{0.62cm}
}]
\end{figure}

\input{sections/0_abstract}    
\input{sections/1_intro}
\input{sections/2-related}
\input{sections/3-method}

\input{sections/4-experiments}

\input{sections/5-conclusion}

\section*{Acknowledgments}
We acknowledge the CINECA award under the ISCRA initiative, for the availability of high-performance computing resources. This work has been supported by the EU Horizon projects ``ELIAS - European Lighthouse of AI for Sustainability'' (No. 101120237) and ``ELSA - European Lighthouse on Safe and Secure AI'' (No. 101070617), and by the PNRRM4C2 project ``FAIR - Future Artificial Intelligence Research'' funded by the EU - NextGenerationEU.

{
    \small
    \bibliographystyle{ieeenat_fullname}
    \bibliography{bibliography}
}

\clearpage
\setcounter{page}{1}
\maketitlesupplementary

\appendix
\input{sections/A-suppl}
\input{sections/B-suppl}

\end{document}

%% file: preamble.tex
\usepackage{booktabs}
\usepackage{array}
\usepackage{multirow}
\usepackage{color}
\usepackage{colortbl}
\usepackage{framed}
\usepackage{bm}
\usepackage{bbm}
\usepackage{xspace}
\usepackage{enumitem}
\usepackage{xparse}
\usepackage{xcolor}
\usepackage{algorithm}
\usepackage{listings}
\usepackage{url}
\usepackage{pifont}

\definecolor{Gray}{gray}{0.93}
\definecolor{LightCyan}{rgb}{0.88,0.95,1}
\definecolor{blond}{rgb}{0.98, 0.94, 0.75}
\definecolor{OurColor}{RGB}{215,237,237}

\definecolor{scale1}{RGB}{234,246,246}
\definecolor{scale2}{RGB}{215,237,237}
\definecolor{scale3}{RGB}{186,224,224}
\definecolor{scale4}{RGB}{133,201,201}
\definecolor{scale5}{RGB}{77,177,177}

\newcommand{\cmark}{\ding{51}}%

\newcommand{\tit}[1]{\smallbreak\noindent\textbf{#1.}}
\newcommand{\tinytit}[1]{\noindent\textbf{#1.}}

\newcommand{\ours}{DICE\xspace}

\newcommand{\datasetDavide}{DICE-D\xspace}

\newcommand\blfootnote[1]{%
  \begingroup
  \renewcommand\thefootnote{}\footnote{#1}%
  \addtocounter{footnote}{-1}%
  \endgroup
}

%% file: sections/0_abstract.tex
\vspace{-32pt}
\begin{abstract}
\vspace{-25pt}

Instruction-based image editing models offer increased personalization opportunities in generative tasks. However, properly evaluating their results is challenging, and most of the existing metrics lag in terms of alignment with human judgment and explainability. To tackle these issues, we introduce \textbf{\ours} (\underline{\textbf{DI}}fference \underline{\textbf{C}}oherence \underline{\textbf{E}}stimator), a model designed to detect localized differences between the original and the edited image and to assess their relevance to the given modification request. \ours consists of two key components: a difference detector and a coherence estimator, both built on an autoregressive Multimodal Large Language Model (MLLM) and trained using a strategy that leverages self-supervision, distillation from inpainting networks, and full supervision. Through extensive experiments, we evaluate each stage of our pipeline, comparing different MLLMs within the proposed framework. We demonstrate that \ours effectively identifies coherent edits, effectively evaluating images generated by different editing models with a strong correlation with human judgment. We publicly release our source code, models, and data at \href{https://aimagelab.github.io/DICE}{project page}. 
\end{abstract}

%% file: sections/1_intro.tex
\vspace{-32pt}
\section{Introduction}
\label{sec:intro}
\blfootnote{$^*$Equal contribution.}

\vspace{-7pt}
The field of Generative AI has recently gained significant attention in both industry and research, driven by the development of models capable of generating content that closely follows the distribution of natural language~\cite{dubey2024llama,guo2025deepseek,achiam2023gpt} and real images~\cite{rombach2022high,podell2023sdxl,esser2024scaling}. Thanks to the availability of cross-modal operators and architectures, multimodal generative approaches, which can seamlessly integrate both the vision and language modalities, are reaching increasing levels of accuracy~\cite{caffagni-etal-2024-revolution,liu2023visual,ye2024mplug,wang2024qwen2}. These advancements are driving the development of new applications and tasks while enhancing user control over the generation process.

Among these proposals, instruction-based image editing models~\cite{brooks2023instructpix2pix, zhang2023magicbrush,geng2024instructdiffusion,huang2024diffusion} enable the modification of an input image based on a free-form textual instruction from the user. This extends traditional text-to-image models, shifting the focus from generating an image from scratch to modifying an existing one while adhering to a given prompt. While such models promise to offer increased personalization levels, the understanding of the results they generate -- and thus their proper evaluation -- clearly becomes more challenging than in traditional text-to-image contexts.

Prior work has approached the evaluation of image editing models by developing annotated benchmarks with ground-truth edited images~\cite{zhang2023magicbrush,shi2020benchmark,shi2021learning} or by proposing metrics based on pre-trained backbones such as CLIP~\cite{radford2021learning} and DINO~\cite{caron2021emerging}. Other methods leverage pre-trained large language models to assess the quality of image edits~\cite{ma2025i2ebench,hui2024hq,betti2023let}. While these proposals might offer a first viable solution to assess the performance of image editing models, they still do not reach a sufficient alignment with human evaluation. Further, being based on either global embedding vectors or replies from pre-trained (and often private) models, explaining their evaluations becomes cumbersome.

Drawing inspiration from these considerations, we take a novel approach and propose a learnable model for understanding image editing outputs, which can provide increased alignment with human preferences and more easily explainable results.
Our approach starts from the hypothesis that modifications made by image editing models typically occur within localized regions of the input image. Accordingly, our model first identifies object-level differences between the original and the edited images and subsequently evaluates the coherence of each detected modification relative to the user's input instruction.
We term our pipeline as \textbf{\ours}, short for \underline{\textbf{DI}}fference \underline{\textbf{C}}oherence \underline{\textbf{E}}stimator. Sample results from our model are shown in Fig.~\ref{fig:first_page}.

Maintaining the goal of creating a fully explainable pipeline, we develop two distinct models, one for each stage, both built on an autoregressive Multimodal Large Language Model (MLLM)~\cite{caffagni-etal-2024-revolution}. While both models leverage the understanding of RoIs, the first primarily focuses on their prediction, whereas the second emphasizes their interpretation to assess coherence with the given prompt. In particular, we first propose a difference detector that, given both images, estimates a set of localized differences which describe the change in content from one image to the other. The model also predicts the category of detected modifications, as either addition, removal, or editing of an object. From a technical standpoint, we train the model in a two-stage protocol, where we first teach the model to identify object-level differences in a self-supervised manner, leveraging pairs of similar images, and then fine-tune it for the specific task using inpainted images. A second model, based on the same architecture, is instead trained to predict the coherence of an object-level modification with respect to the modification request made by the user and provide a textual rationale for each decision made.

We experimentally assess the appropriateness of our strategy by carefully investigating the performance of both our difference detector and our coherence estimator, in comparison with existing solutions and when employing different MLLMs. Further, we conduct a dedicated user study to measure the alignment of the evaluations predicted by our proposal in terms of model ranking and evaluation. Experimental results demonstrate that our proposal achieves increased ranking capabilities while benefiting from an explainable-by-default approach. Further, when integrated in existing metrics, it acts as the basis for building model evaluation metrics with increased human alignment.

%% file: sections/2-related.tex
\section{Related Work}
\label{sec:related}

\tinytit{Image Editing Models}
Recent developments in Generative Adversarial Networks (GANs)~\cite{goodfellow2014generative,karras2019style} and diffusion models~\cite{rombach2022high,podell2023sdxl,esser2024scaling,dhariwal2021diffusion} have driven the rise of AI-generated content. However, users increasingly demand not only the creation of new visual data but also the modification of existing images, altering properties such as style~\cite{karras2019style,gatys2016image} and content~\cite{brooks2023instructpix2pix,fu2024guiding}. 
Within this domain, tasks vary based on the input used for conditioning. Image inpainting~\cite{suvorov2022resolution,chiu2024brush2prompt}, for instance, uses user-designed masks to guide edits in specific regions. In contrast, our work focuses on instruction-based image editing, where the model relies solely on textual instructions and the original image.

Within instruction-based editing, the most traditional approaches rely on diffusion models fine-tuned for instruction following. A fundamental contribution in this domain is InstructPix2Pix~\cite{brooks2023instructpix2pix} which is trained on fully synthetic triplets composed of an input image, an editing instruction, and the resulting edited image.
However, reliance on synthetic data may introduce noise into the predictions. To mitigate this limitation, MagicBrush~\cite{zhang2023magicbrush} refines the InstructPix2Pix framework using a high-quality, curated dataset. Differently, InstructDiffusion~\cite{geng2024instructdiffusion} is a Stable Diffusion-based model that is trained to perform a wide variety of tasks including image editing, semantic segmentation, and object detection.
Further, HIVE~\cite{zhang2024hive} proposes a reward-based training procedure that incorporates human feedback to enhance its instruction following capabilities. 

A complementary research direction involves integrating an MLLM to improve language understanding within the editing process. In this regard, MGIE~\cite{fu2024guiding} employs a frozen MLLM to generate concise and representative edit instructions that guide the diffusion model during image editing, whereas SmartEdit~\cite{huang2024smartedit} captures the interactions between image and text by extracting salient information from the image tokens produced by an MLLM.

\begin{figure*}[t]
    \centering
    \includegraphics[width=0.99\linewidth]{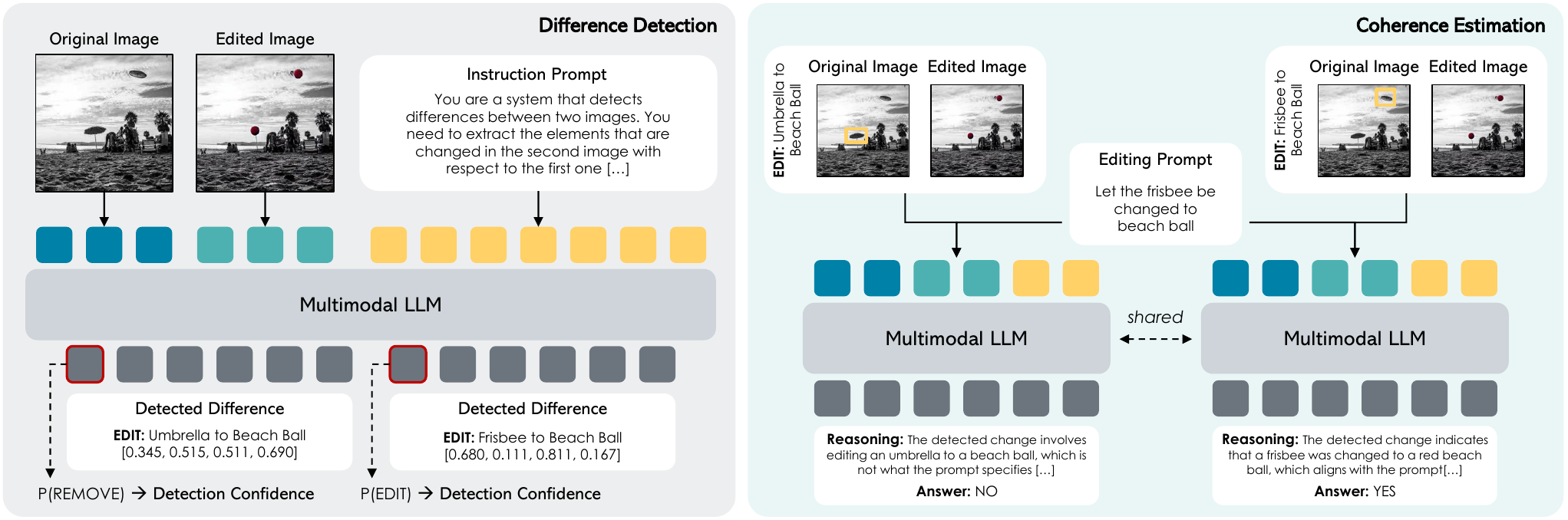}
    \vspace{-0.15cm}
    \caption{
    Illustration of \ours. We employ an MLLM and fine-tune it for two different tasks. In the first stage (difference detection), the MLLM is trained to detect semantic differences between the original image and the edited one. In the second stage (coherence estimation), the MLLM is instructed to analyze and assess the coherence of each detected difference with respect to the given user prompt.}
    \vspace{-0.3cm}
    \label{fig:method}
\end{figure*}
\tit{Image Editing Evaluation and Benchmarks}
Instruction-based image editing models have traditionally been evaluated along two dimensions: prompt adherence and background preservation. The former assesses the extent to which the output aligns with the given prompt instruction, whereas the latter evaluates the preservation of the unmodified contextual information. While benchmarks featuring annotated ground-truth edited images 
are available~\cite{zhang2023magicbrush,shi2020benchmark,shi2021learning}, our work prioritizes a reference-free evaluation that eliminates the need for human-defined labels.
In this setting, background preservation is typically evaluated by using either CLIP~\cite{radford2021learning} or DINO~\cite{caron2021emerging} to compute cosine similarities between the visual tokens extracted from the source and edited images~\cite{brooks2023instructpix2pix,zhang2024hive,fu2024guiding}.
Likewise, prompt adherence is usually quantified using CLIP-based similarities between the target caption and the edited image~\cite{sheynin2024emu,zhang2023magicbrush}.

With the recent proposal of MLLMs, these methods have been employed for prompt adherence evaluation. For instance, in HQ-Edit~\cite{hui2024hq}, GPT-4V~\cite{achiam2023gpt} is used to assess coherence and alignment metrics.
Additionally, I$^2$E Bench~\cite{ma2025i2ebench} incorporates human annotation for certain queries, subsequently leveraging GPT-4V for question answering.
However, reliance on GPT-based evaluations introduces potential variability and reproducibility issues, as model updates may alter the final scores. In contrast, we propose an interpretable solution based on an open-source MLLM, thereby increasing both the rationale behind our scoring mechanism and the reproducibility of the results.

\tit{Multimodal Large Language Models}
With the introduction of LLMs, the research community has directed great effort towards connecting visual modalities with textual reasoning~\cite{liu2023visual,cocchi2025llava,bucciarelli2024personalizing}. Building on these efforts, MLLMs have also been used to address visual grounding tasks. Among these, Shikra~\cite{chen2023shikra} leverages a novel training paradigm to give the MLLM the capacity to reason with bounding boxes, making it able to precisely locate objects and relations in the image without relying on external components. 
LION~\cite{chen2024lion} improves spatial awareness through progressive integration of spatial knowledge and soft prompting. Groma~\cite{ma2024groma} decomposes images into region tokens, enhancing localization via user instructions. In contrast, LISA~\cite{lai2024lisa} and GROUNDHOG~\cite{zhang2024groundhog} perform segmentation incorporating pixel-level representations for fine-grained visual understanding.

Compared to existing MLLMs for visual grounding, our model leverages multiple images to detect semantic differences, introducing a novel setting that requires an MLLM capable of multi-image understanding.
In this context, Idefics3~\cite{laurenccon2024building} is able to comprehend multiple visual inputs interleaved with text, using a novel image encoding pipeline. 
Differently, Qwen~\cite{wang2024qwen2} extends classical multimodal capabilities to video and multi-image understanding thanks to a novel encoding strategy and positional embedding technique for the visual inputs that takes into account their spatial dimensions. Finally, mPLUG-Owl3 \cite{ye2024mplug} introduces a novel attention-based strategy to efficiently encode long image sequences. 
Building on these multi-image models, we extend the capabilities of MLLMs by introducing a novel framework for multi-image detection, enabling the identification of semantic differences across multiple images within a unified system.

%% file: sections/3-method.tex
\section{Proposed Method}
\label{sec:method}

\subsection{Preliminaries}
The objective of our approach is to assess the quality of outputs generated by instruction-based image editing models~\cite{brooks2023instructpix2pix,zhang2024hive,fu2024guiding}. 
Formally, given an original image $\bm{x}$ and a textual instruction $\bm{t}$, an image editing method $f$ generates a new image $\bm{e} = f(\bm{x}, \bm{t})$, which incorporates the modifications specified by the textual prompt $\bm{t}$, while preserving the contextual integrity of $\bm{x}$.

While the modifications introduced by image editing models can be diverse -- owing to the expressiveness of textual instructions and the flexibility of generative models -- we focus specifically on object-level modifications. By means of this choice, we consider transformations that are clearly localizable inside of the input scene. In particular, we assume that the original image contains a finite set of objects, denoted as $\{ o_i \}_{i=1}^{n}$, and that the textual instruction $\bm{t}$ specifies an operation to be applied to it. Given a correct application of the specified operation, the expected edited image, denoted as $\tilde{\bm{e}}$, serves as the ground-truth for evaluating the intended modification.

\tit{Object-level Modifications}
Without loss of generality, we assume that the alterations required in the textual prompt can be classified as either the addition, removal, or modification of objects inside the input image. Specifically, we consider three categories of alterations, namely:
\begin{enumerate}
    \item \textbf{Addition of an object.} A new object instance $o_{n+1}$ is introduced into the scene, yielding a new image representation. Treating images as a set of objects, and with a slight abuse of notation, this operation results in a new image that can be expressed as
    $
    \tilde{e} = \left( \bigcup_{i=1}^{n} o_i \right) \cup \{ o_{n+1} \}.
    $
    \item \textbf{Removal of an object.} The operation removes an object $o_j$ which was present in the original image, which results in 
    $
    \tilde{e} = \left( \bigcup_{i=1}^{j-1} o_i \cup \bigcup_{i=j+1}^{n} o_i \right).
    $
    \item \textbf{Edit of an object.} An existing object $o_j$ is altered in at least one visual aspect. Let $\tilde{o}_j$ denote the modified object; the updated image is then represented as
    $
    \tilde{e} = \left( \bigcup_{i=1}^{n} o_i \setminus \{ o_j \} \right) \cup \{ \tilde{o}_j \}.
    $
\end{enumerate}

\subsection{Overview of the Approach}
An image editing network can produce an image $\bm{e}$ that does not accurately respect all the object-level operations required to produce $\tilde{e}$. To quantify this discrepancy, we propose \underline{\textbf{DI}}fference \underline{\textbf{C}}oherence \underline{\textbf{E}}stimator (\textbf{\ours}): a two-step approach in which (i) we detect the object-level differences between the edited and the original image and then (ii) we evaluate the coherence of each difference with the textual prompt to determine whether a discrepancy reflects the intended modifications or constitutes an unwanted alteration. An overview of the two stages of \ours is depicted in Fig.~\ref{fig:method}.
\tit{Difference Detection} 
In this stage, we extract the set of objects that are present in the original image $\bm{x}$ but absent or changed in the predicted image $\bm{e}$ or vice versa. We denote this collection of differences as \textit{semantic difference}, which is formally defined as
\begin{equation}
    \Delta(\bm{x},\bm{e}) = \{ o \mid o \in \{\bm{x} \setminus \bm{e}\} \cup \{\bm{e} \setminus \bm{x}\} \}.
\end{equation}
The aforementioned set of differences is estimated through a learnable difference detector $D(\bm{x},\bm{e})$ that, given the original and modified image, predicts a set of object-level differences which serves as an approximation of $\Delta(\bm{x},\bm{e})$, expressed in terms of RoIs. Notably, this detection process is executed independently of the textual prompt $\bm{t}$ to ensure that the outcome is not biased by the intended edit.

\tit{Coherence Estimation}
In the second step, we evaluate each detected difference individually to assess its coherence to the textual prompt. Specifically, a coherence estimator operates on each detected object-level change $o_i \in D(\bm{x}, \bm{e})$, determining whether the visual modifications align with the intended edit $\tilde{\bm{e}}$. Formally,
\begin{equation}
    C(\bm{x}, \bm{e}, \bm{t}, o_i) = 
    \begin{cases}
    1 & \text{if } o_i \in \tilde{\bm{e}}, \\
    0 & \text{otherwise}.
    \end{cases}
    \label{eq:coherence}
\end{equation}
This process enables the differentiation between intended and unintended edits, thereby facilitating a comprehensive and interpretable evaluation of the task.

\subsection{Detecting Differences}
To build the difference detector and the coherence estimator, we design two novel architectures building upon the general structure of autoregressive Multimodal Large Language Models (MLLMs), which offer a sufficiently general framework for encoding images, prompts, and RoIs.

For difference detection, we train an MLLM $D(\mathbf{x},\mathbf{e})$ to predict object-level differences between two images in terms of (i) the type of modification (\ie, addition, removal, edit), (ii) the RoI of the modified region, and (iii) the subjects involved in the modification. Formally, the difference detector operates as:
\begin{equation}
    D(\bm{x}, \bm{e}) = \left[ (c_i, S_i, b_i) \right]_{i=1}^{k},
\end{equation}
where $c_i$ denotes a command from $\{\text{\texttt{ADD}}, \text{\texttt{EDIT}}, \text{\texttt{REMOVE}}\}$, $S_i$ describes the modified object in free text, and $b_i \in \mathbb{R}^4$ defines the bounding box coordinates in the format $(x_{\text{min}}, y_{\text{min}}, x_{\text{max}}, y_{\text{max}})$.

Each tuple $(c_i, S_i, b_i)$ is fully specified in the textual domain, and the model is optimized to output structured text representations of modifications, formatted as: \texttt{"COMMAND: object, [bb1, bb2, bb3, bb4]"}. This structured output ensures precise localization and identification of changes.

To estimate the confidence of a predicted RoI, we compute the probability of the predicted command $c_i$ relative to the total probability mass of all possible commands. This confidence measure aligns with conventional object detection confidence estimation techniques~\cite{carion2020end,Girshick_2015_ICCV}.

\tit{Training Procedure Overview}
\label{sec:difference_detection_training}
Our training pipeline consists of two phases: (i) a pre-training stage leveraging real image pairs and (ii) a fine-tuning stage using images modified with inpainting models. 
Notably, the model is intentionally not trained on instruction-based paired edited images due to two key considerations. First, such training may lead to overfitting to specific editing architectures while underperforming on unseen methods. Second, instruction-edited image pairs would necessitate extensive human annotations for commands and RoIs, introducing significant cost and time constraints.

\tit{Stage 1: Learning to Compare Similar Images}
In this stage, we instruct the difference detector to identify object-level differences when considering pairs of similar real images. Specifically, the MLLM is trained to predict objects that are not mutually present in both images, leveraging an existing object detection dataset.

In detail, we employ images and object-level annotations from the LVIS dataset~\cite{gupta2019lvis}. We identify pairs of similar images by computing the cosine similarity in the DINOv2~\cite{oquab2023dinov2} embedding space, and retaining only those with a cosine similarity greater than 0.6, ensuring high visual similarity. In addition, we make sure that the aforementioned pairs contain at least one common object class and differ by fewer than 15 object classes. This filtering results in a total of 118k image pairs.
When encoding image pairs in this stage, objects present in the first image but absent in the second are labeled as \texttt{REMOVE}, while objects appearing in the second image but not in the first are labeled as \texttt{ADD}. Objects exhibiting an intersection-over-union above a predefined threshold across both images are classified as \texttt{EDIT}. Further details regarding dataset construction are provided in the supplementary material.

\tit{Stage 2: Learning to Detect Inpainted Areas}
In the second stage, we refine the model by bridging the gap between pre-training on real image pairs and application to edited images.  To accomplish this, we select images from the LVIS dataset and sample non-overlapping annotated objects. Starting from LVIS images, we employ inpainting to generate original and edited image pairs. For each object, an operation is randomly selected among $\{\texttt{ADD}, \texttt{EDIT}, \texttt{REMOVE} \}$. Inpainting is then performed using LaMa~\cite{suvorov2022resolution}: for the added objects, the corresponding region is inpainted on the original image, while for removed objects the inpaint is performed on the edited one. For the edit operation, 30\% of the edited objects undergo a color change, while the remaining 70\% are substituted with a different object. We rely on GPT-4o to generate target objects suitable for substitution by prompting it with a caption of the original image as well as the original object. To perform the edits, the Kandinsky 2.2 inpaint model~\cite{razzhigaev2023kandinsky} is employed to inpaint selected regions. This process leads to a training set of 97k elements and a test set of 19k.

\subsection{Estimating the Coherence}
\label{sec:coherence_training}
Given the original and edited image, along with a detected difference, the coherence estimator produces a binary decision (\ie, Yes/No) and a textual rationale indicating whether the modification is coherent with the intended edit. To induce the model to focus on a specific difference, the bounding box of the detected difference is visually depicted on the original and edited image before they are fed to the coherence estimator. Further, we also include the type of modification $c_i$ and its textual description $S_i$ as input to the model.

To train the model, we annotated 116 samples from the EmuEdit~\cite{sheynin2024emu} test set, with the corresponding edited images generated using InstructDiffusion~\cite{geng2024instructdiffusion}. For each sample, we manually annotated both the observed differences and a binary coherence indicator. In addition, we employed GPT-4o to generate a rationale for each annotation, which was subsequently refined through human review to eliminate any inaccuracies. While the command $c$ and the subjects $S$ were provided in textual format, the bounding boxes were directly depicted on the images using a color-coded scheme: red for additions, green for edits, and blue for removals. Specifically, additions are superimposed on the edited image, whereas the bounding boxes corresponding to removals and edits are applied solely to the original image.

%% file: sections/4-experiments.tex
\section{Experimental Results}
\label{sec:experiments}

\input{Tables/table_1_BBox_Comparison}

\input{Tables/table_2_coherence_eval}

\subsection{Implementation and Training Details}
In our experiments, we consider three recent MLLMs designed to support multi-image input tasks: Idefics3-8B~\cite{laurenccon2024building}, Qwen2-VL-7B~\cite{wang2024qwen2}, and mPLUG-Owl3-7B~\cite{ye2024mplug}. Following the training steps described in Sec.~\ref{sec:difference_detection_training} and Sec.~\ref{sec:coherence_training}, we fine-tune each model using QLoRA~\cite{dettmers2023qlora}. Specifically, the vision encoder and projector of each model remain frozen, while the language model is quantized and augmented with LoRA~\cite{hu2022lora} adapters. LoRA is applied to the self-attention and MLP layers across all 32 Transformer blocks.

During the second stage of the difference detector component, we observed that the model may be susceptible to artifacts introduced by the inpainting model, which can compromise its generalizability across different editing outputs. To remove the effect of these low-level imprints, we apply random JPEG compression at 15\% and 50\% magnitude, following an established practice in deepfake detection~\cite{wang2020cnn}. Further details regarding model configuration and training are provided in the supplementary material.

\subsection{Evaluating Detected Differences}
\label{sec:evaldifferences}
To evaluate the performance of our detection and coherence pipeline, we construct a new dataset, termed as \datasetDavide, specifically designed for the task. This dataset comprises 800 edited images extracted from the I$^2$EBench~\cite{ma2025i2ebench} benchmark. We select images and prompts from a subset of the edit categories that involve object-based modifications, in particular: color alteration, counting (only if involving a single object change), object removal, and object replacement. Each pair of images is manually annotated with detected differences represented as bounding boxes, editing commands (\ie, \texttt{ADD}, \texttt{REMOVE}, \texttt{EDIT}), and textual descriptions. Also, we assign a coherence label for each bounding box to indicate its alignment with the prompt.

\tit{Difference Detection Results}
We first evaluate the ability of the difference detection model to detect modifications between the original and edited images. Following object detection literature~\cite{lin2014microsoft,carion2020end}, we compute the mean average precision ($\mathsf{AP}$) between predicted and labeled bounding boxes from \datasetDavide. Specifically, we average results at 10 IoU thresholds from 0.5 to 0.95.
Further, we include $\mathsf{AP_{50}}$ (average precision at 50\% IoU),  $\mathsf{AP_{75}}$ (average precision at 75\% IoU), $\mathsf{AP_M}$ (for medium-sized objects), and $\mathsf{AP_L}$ (for large-sized objects), ensuring a comprehensive evaluation across different object scales and IoU thresholds.

Results are reported in Table~\ref{tab:detection_results}, where we consider both class-agnostic detection and class-aware detection. In class-agnostic detection, all predictions are treated as belonging to the same category, to evaluate only the bounding box location. Vice versa, in class-aware detection, both the bounding box location and the editing command are considered, evaluating the accuracy of both the detection and the associated command.
As shown, using Idefics as the base model consistently achieves the best performance across all metrics, with an increase of 17.1 and 20.0 in $\mathsf{AP}$ in a class-agnostic setting compared to Qwen and mPLUG. Similarly, when considering the command labels, Idefics outperforms Qwen and mPLUG by 11.8 and 13.9. This superior performance may be attributed to the distinctive image encoding pipeline employed by Idefics, in which each image is resized to a larger dimension and segmented into a grid of crops, with each crop being encoded independently. Although this approach increases the amount of visual information processed, the grid-based encoding simultaneously introduces a localization factor to the captured visual details that can be leveraged in a task of detection.

When validating the contribution of the key components of our pipeline, it is worth noting that the initial pre-training stage significantly contributes to the final performance. Specifically, training Idefics in both stages results in improvements of 1.6 and 1.2 in class-agnostic and class-aware $\mathsf{AP}$, respectively, compared to training solely on the second stage dataset. Notably, even training exclusively on the first stage enhances class-agnostic detection compared to Qwen (\ie, +0.4 $\mathsf{AP}$). This observation implies that predicting missing objects in pairs of similar real images not only facilitates the learning of prediction syntax but also enables the transfer of relevant knowledge to the difference detection task in edited images.

Moreover, incorporating the extracted confidence score enhances difference localization results, with improvements of 5.5 and 3.1 $\mathsf{AP}$ observed for Idefics-based models trained on both stages and solely on the second stage respectively, compared to using a fixed confidence value of 1. Additionally, using random confidence values results in worse performance, further indicating that the uncertainty in the command correlates with the actual prediction confidence.

\input{Tables/table_3_editing_models}

\tit{Coherence Estimation Results}
The coherence estimation model assesses whether each detected difference is correctly aligned with the prompt using a binary prediction (\ie, Yes/No). Moreover, the coherence step incorporates explicit model reasoning~\cite{wei2022chainofthoughts}, thereby enhancing the interpretability of the results. Under this setting, models are evaluated on \datasetDavide, in terms of coherence over ground-truth areas and coherence over detected areas. To measure the coherence accuracy, ground-truth differences are fed into the coherence estimator, and results are evaluated against the manually annotated coherence. Instead, for the coherence over detected areas, the performance of the entire \ours pipeline is measured through $\mathsf{AP}$, considering the coherence label of the difference as the ground-truth category of the bounding box. This is done by extracting differences with the detector and subsequently predicting the coherence of each difference.

Results are summarized in Table~\ref{tab:coherence_results}, evaluating the effectiveness of various MLLMs as coherence estimators. Idefics consistently achieves the highest overall performance on $\mathsf{AP}$, surpassing mPLUG and Qwen by 11.0 and 12.7 points, respectively, while maintaining comparable coherence accuracy. Additionally, incorporating extracted confidence into the $\mathsf{AP}$ computation leads to performance improvements across all evaluated MLLMs. This highlights the importance of confidence extraction in the difference detection phase and motivates further research on aligning token probability with confidence estimation in object detection. Based on these results, Idefics3-8B is selected as the base MLLM for \ours in all subsequent experiments.

\subsection{Evaluation of Image Editing Models}
\tinytit{User Study Details} 
We conduct a user study to evaluate our proposed framework within the context of instruction-based image editing. In detail, we collect 200 prompts along with their corresponding original images, 50\% from MagicBrush~\cite{zhang2023magicbrush} and 50\% from I$^2$EBench~\cite{ma2025i2ebench}. For MagicBrush, we use GPT-4o to filter out prompts that do not involve object modification. Differently, for the I$^2$EBench subset, we employ the same selection criteria as in the construction of \datasetDavide (cf. Sec.~\ref{sec:evaldifferences}). We select five state-of-the-art generators and generate 200 edited images per generator. Specifically, our model selection includes HIVE~\cite{zhang2024hive}, InstructPix2Pix~\cite{brooks2023instructpix2pix}, MGIE~\cite{fu2024guiding}, MagicBrush~\cite{zhang2023magicbrush}, and InstructDiffusion~\cite{geng2024instructdiffusion}. 
To assess the quality of the generated edits, we asked 10 human annotators to evaluate each sample on two dimensions: \textit{prompt adherence} and \textit{background preservation}. Prompt adherence measures how well the modifications in the edited image align with the instructions given in the prompt. Differently, background preservation assesses how well the elements that were not intended to be modified remained unchanged. Ratings for each dimension were recorded on a 5-point Likert scale. We refer the user to the supplementary materials for more detailed information about the user study.

\tit{Model Editing Ranking}
To evaluate the performance of editing models, we establish three axes to rank the considered generators using \ours. \textit{Correct edits} measure the percentage of images in which at least one detected difference is classified as coherent. Conversely, \textit{unwanted edits} quantify the percentage of the total image area containing differences marked as non-coherent. Lastly, \textit{no visual change} represents the percentage of images where \ours does not detect any visual alteration, potentially due to insufficient prompt clarity, lack of understanding, or inability of the model to execute the requested modification. Notably, \textit{correct} and \textit{unwanted edits} can be measured with the average prompt adherence and background preservation ratings from the user study. Similarly, from the user study, \textit{no visual change} is assessed by identifying images rated 5 in background preservation and 1 in prompt following.

In Table~\ref{tab:benchmark}, we compare the rankings generated by our evaluation with those obtained from the user study. As it can be seen, model rankings from \ours align with human evaluations. In particular, InstructDiffusion and MagicBrush emerge as the top-performing models both according to \ours and human ratings, excelling in performing correct modifications while effectively preserving the background. Likewise, the ranking order for \textit{no visual change} remains consistent across \ours and the user study, further validating the reliability of our approach.

\begin{figure*}[t]
    \centering
    \includegraphics[width=0.98\linewidth]{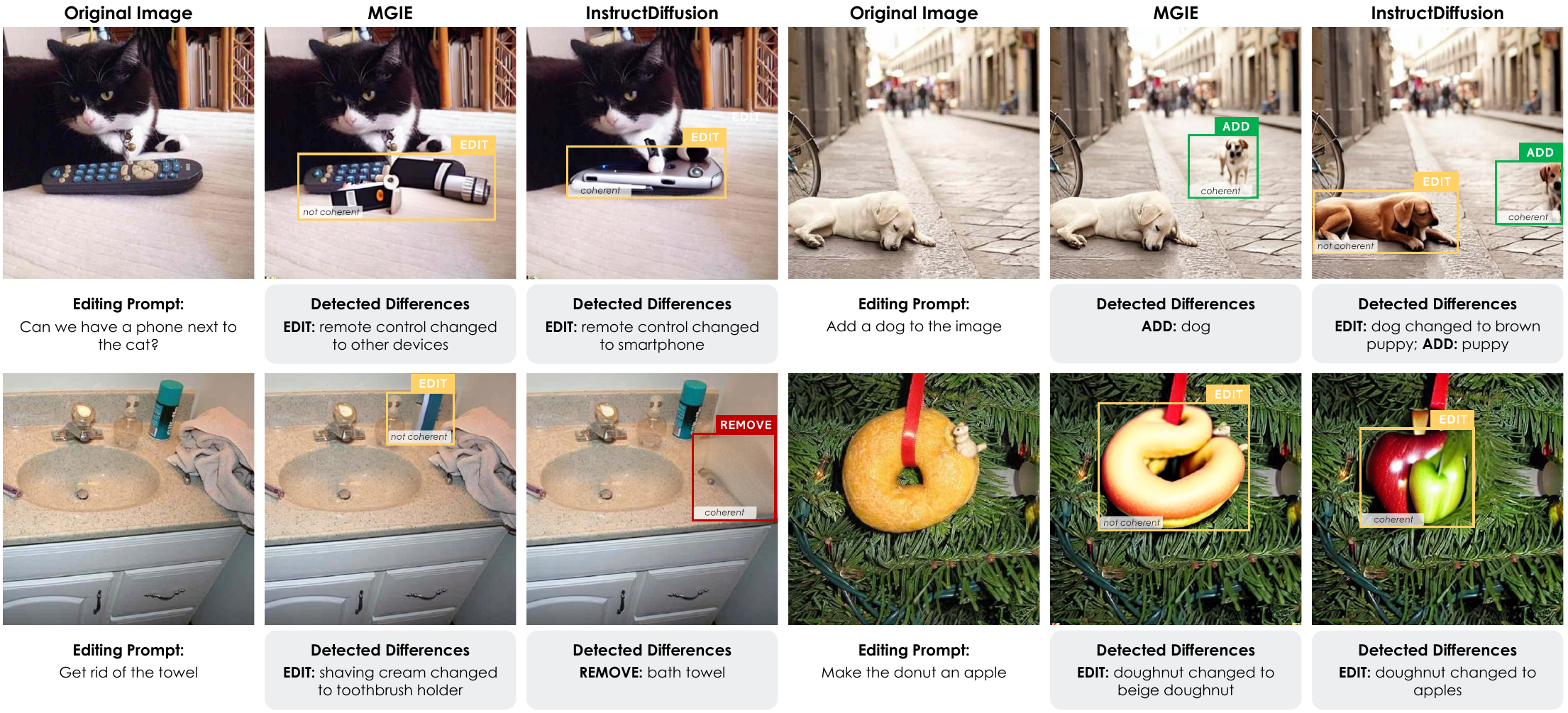}
    \vspace{-0.15cm}
    \caption{Qualitative samples of \ours applied on images edited by MGIE~\cite{fu2024guiding} and InstructDiffusion~\cite{geng2024instructdiffusion} models.}
    \vspace{-0.4cm}
    \label{fig:qualitatives}
\end{figure*}

To qualitatively validate the predictions of \ours, Fig.~\ref{fig:qualitatives} presents editing samples showcasing difference detection and coherence predictions across various prompts and models. Notably, \ours effectively identifies image differences and evaluates their coherence with the editing prompt.

\tit{Correlation with Human Judgment}
To further validate the alignment between the detected differences and human judgment, we integrate \ours predictions within CLIP-based evaluation metrics measuring image similarity (CLIP-I) for background preservation and image-text similarity (CLIP-T) for prompt adherence~\cite{hessel2021clipscore}. In particular, CLIP-I measures the similarity between the original and the edited image. To compute CLIP-T, instead, we caption the original image using an MLLM (\ie, Idefics3-8B in our experiments) and feed the extracted caption along with the editing prompt to GPT-4o to produce a caption of the ideal edited image (\ie, target caption). Through CLIP-T we then measure the CLIP similarity between the target caption and the actual edited image.
To integrate \ours in CLIP-I and CLIP-T, we modify both the original and edited images by selectively masking specific regions, taking into account \ours predictions. Specifically, for background preservation, we mask coherent differences, whereas, for prompt adherence assessment, non-coherent differences are patched. 

\input{Tables/table_4_correlation_scores}

For this experiment, we compare the correlation scores of standard CLIP-I and CLIP-T metrics with those obtained by masking original and edited images using \ours. Results are shown in Table~\ref{tab:correlation}, where we measure Spearman's $\rho$, Pearson's $\rho_s$, and Kendall's $\tau$ correlation coefficients between automated evaluation scores and human judgment. We also compare when randomly masking areas from the differences detected by \ours, or when masking all detected differences, including both coherent and non-coherent changes.
As it can be seen, \ours effectively enhances the correlation of both CLIP-I and CLIP-T metrics with human ratings. Specifically, for background preservation, CLIP-I exhibits the strongest correlation with human judgments when masking only coherent differences, achieving a Pearson’s score of 54.5. This is a significant improvement over the baseline score of 51.1 (no masking), and substantially higher than the scores obtained with random masking or masking all detected differences. This result suggests that excluding correctly modified regions allows CLIP-I to better focus on the actual background, producing a metric that more accurately aligns with human perception.

For prompt adherence, CLIP-T shows the strongest correlation when we patch non-coherent differences, achieving a Pearson’s score of 24.6, compared to 21.5 without masking. Conversely, masking all detected differences leads to a drop to 13.3, likely due to indiscriminate removal of both erroneous and correct edits. This outlines that selectively removing only incorrect modifications helps CLIP-T focus on prompt-relevant changes, thereby improving alignment with human evaluations.

Overall, these results demonstrate that \ours enhances the reliability of CLIP-based metrics, making them more aligned with human perception of editing quality. In particular, distinguishing between coherent and non-coherent changes is crucial for obtaining meaningful evaluations.

%% file: Tables/table_1_BBox_Comparison.tex
\begin{table*}[t]
  \centering
  \setlength{\tabcolsep}{.35em}
  \resizebox{\linewidth}{!}{
  \begin{tabular}{lclc cccc c ccccc c cccccccc}
    \toprule
    & & & & & & \multicolumn{2}{c}{\textbf{Training}} & & \multicolumn{5}{c}{\textbf{Class-agnostic Detection}} & & \multicolumn{8}{c}{\textbf{Class-aware Detection}} \\
    \cmidrule{7-8} \cmidrule{10-14} \cmidrule{16-23}
    & & & & \textbf{Confidence} & & Stage 1 & Stage 2 & & $\mathsf{AP}$ & $\mathsf{AP_{50}}$ & $\mathsf{AP_{75}}$ & $\mathsf{AP_M}$ & $\mathsf{AP_L}$ & & $\mathsf{AP}$ & $\mathsf{AP_{50}}$ & $\mathsf{AP_{75}}$ & $\mathsf{AP_M}$ & $\mathsf{AP_L}$ & $\mathsf{AP_{ADD}}$ & $\mathsf{AP_{REM}}$ & $\mathsf{AP_{EDIT}}$ \\
    \midrule
    \rowcolor{Gray}
    \multicolumn{23}{l}{\textit{Alternative MLLMs}}\\
    & & Qwen2-VL-7B~\cite{wang2024qwen2} & & - & & \cmark & \cmark & & 2.0 & 5.3 & 1.3 & 0.3 & 3.4 & & 1.6 & 3.3 & 1.2 & 0.1 & 2.4 & 0.1 & 0.1 & 3.2 \\
    & & Qwen2-VL-7B~\cite{wang2024qwen2} & & \cmark & & \cmark & \cmark & & 2.3 & 5.6 & 1.8 & 0.1 & 3.5 & & 1.6 & 4.1 & 1.1 & 0 & 1.6 & 0.1 & 0.1 & 3.2 \\
    \cmidrule{3-23}
    & & mPLUG-Owl3-7B~\cite{ye2024mplug} & & - & & \cmark & \cmark & & 3.8 & 10.3 & 2.4 & 0.5 & 5.1 & & 2.5 & 6.7 & 1.7 & 0.3 & 3.3 & 1.1 & 1.5 & 5.1 \\
    & & mPLUG-Owl3-7B~\cite{ye2024mplug} & & \cmark & & \cmark & \cmark & & 5.2 & 13.7 & 3.2 & 0.5 & 7.3 & & 4.4 & 10.6 & 3.1 & 0.3 & 5.5 & 0.1 & 7.3 & 4.8 \\
    \midrule
    \rowcolor{Gray}
    \multicolumn{23}{l}{\textit{Ablation Studies}}\\
    & & Idefics3-8B~\cite{laurenccon2024building} & & - & & - & \cmark & & 15.1 & 30.1 & 13.4 & 8.5 & 18.6 & & 9.4 & 18.1 & 9.3 & 5.5 & 11.0 & 7.3 & 7.5 & 13.4 \\
    & & Idefics3-8B~\cite{laurenccon2024building} & & - & & \cmark & \cmark & & 16.7 & 30.7 & 16.0 & 10.4 & 20.3 & & 10.6 & 17.9 & 10.9 & 6.6 & 12.5 & 7.4 & 7.7 & 16.5 \\
    & & Idefics3-8B~\cite{laurenccon2024building} & & \hspace{0.44cm}\cmark$_\text{rand}$ & & \cmark & \cmark & & 16.8 & 31.2 & 16.2 & 10.3 & 20.5 & & 9.7 & 16.5 & 10.0 & 6.7 & 11.2 & 7.0 & 7.9 & 14.2 \\
    \cmidrule{3-23}
    & & Idefics3-8B~\cite{laurenccon2024building} & & \cmark & & \cmark & - & & 2.9 & 5.1 & 2.8 & 1.9 & 4.2 & & 1.4 & 2.2 & 1.4 & 0.9 & 1.8 & 2.0 & 0.4 & 1.8 \\
    & & Idefics3-8B~\cite{laurenccon2024building} & & \cmark & & - & \cmark & & 18.2 & 36.8 & 16.0 & 9.8 & 22.7 & & 12.2 & 22.2 & 12.1 & 7.3 & 15.3 & 7.2 & 13.4 & 15.8 \\
    \midrule
    \rowcolor{Gray}
    \multicolumn{23}{l}{\textbf{DICE (Ours)}}\\
    \rowcolor{OurColor}
    & & Idefics3-8B~\cite{laurenccon2024building} & & \cmark & & \cmark & \cmark & & \textbf{22.3} & \textbf{40.1} & \textbf{21.9} & \textbf{13.5} & \textbf{36.8} & & \textbf{15.5} & \textbf{24.8} & \textbf{16.5} & \textbf{11.4} & \textbf{15.4} & \textbf{19.6} & \textbf{14.8} & \textbf{16.4} \\
    \bottomrule
  \end{tabular}
      }
  \vspace{-0.15cm}
  \caption{Performance comparison of various MLLMs in the difference detection stage of our pipeline, evaluated under both class-agnostic and class-aware settings. Results are presented in terms of $\mathsf{AP}$ metrics across various training configurations.}
  \label{tab:detection_results}
  \vspace{-0.35cm}
\end{table*}

%% file: Tables/table_2_coherence_eval.tex
\begin{table}[t]
  \centering
  \setlength{\tabcolsep}{.22em}
  \resizebox{\linewidth}{!}{
  \begin{tabular}{cclc ccc c cccc}
    \toprule
     & & & & & & \textbf{Coherence over} & & \multicolumn{3}{c}{\textbf{Coherence over}} \\
    & & & & & & \textbf{GT Areas} & & \multicolumn{3}{c}{\textbf{Detected Areas}} \\
    \cmidrule{7-7} \cmidrule{9-11}
     & & & & \textbf{Confidence} & & $\mathsf{Accuracy}$ & & $\mathsf{AP}$ & $\mathsf{AP_{50}}$ & $\mathsf{AP_{75}}$ \\
    \midrule
    \rowcolor{Gray}
    \multicolumn{11}{l}{\textit{Alternative MLLMs}} \\
    & &  Qwen2-VL-7B~\cite{wang2024qwen2} & & - & & 76.6 & & 1.4 & 3.8 & 0.6  \\
    & & Qwen2-VL-7B~\cite{wang2024qwen2}& & \cmark & & 76.6 & & 1.8 & 4.9 & 0.9  \\
    \cmidrule{3-11}
    & & mPLUG-Owl3-7B~\cite{ye2024mplug} & & - & & 85.9 & & 3.8 & 10.0 & 2.4  \\
    & & mPLUG-Owl3-7B~\cite{ye2024mplug} & & \cmark & & 85.9 & & 4.5 & 11.3 & 2.9  \\
    \midrule
    \rowcolor{Gray}
    \multicolumn{11}{l}{\textbf{\ours (Ours)}} \\
    & & Idefics3-8B~\cite{laurenccon2024building} & & - & & 85.4 & & 12.0 & 20.3 & 12.0  \\
    \rowcolor{OurColor}
     & & Idefics3-8B~\cite{laurenccon2024building} & & \cmark & & 85.4 & & \textbf{15.5} & \textbf{26.0} & \textbf{16.1}  \\
    \bottomrule
  \end{tabular}
  }
  \vspace{-0.15cm}
  \caption{Performance comparison of various MLLMs in the coherence estimation stage of our pipeline. Coherence accuracy is measured using ground-truth differences, while coherence over detected areas employs the output of the first stage of our pipeline and is evaluated using $\mathsf{AP}$ metrics.}
  \label{tab:coherence_results}
  \vspace{-0.4cm}
\end{table}

%% file: Tables/table_3_editing_models.tex
\begin{table*}[t]
    \centering
    \setlength{\tabcolsep}{.3em}
    \resizebox{0.97\linewidth}{!}{
    \begin{tabular}{lc c >{\centering\arraybackslash}m{2.4cm}c>{\centering\arraybackslash}m{2.4cm} c >{\centering\arraybackslash}m{2.4cm}c>{\centering\arraybackslash}m{2.4cm} c >{\centering\arraybackslash}m{2.4cm}c>{\centering\arraybackslash}m{2.4cm}}
        \toprule
        & & & \multicolumn{3}{c}{\textbf{Correct Edits}}
        & & \multicolumn{3}{c}{\textbf{Unwanted Edits}}
        & & \multicolumn{3}{c}{\textbf{No Visual Change}} \\
        \cmidrule(lr){4-6} \cmidrule(lr){8-10} \cmidrule(lr){12-14}
            \textbf{Editing Models} & & & \textbf{\ours (\%)} $\uparrow$ & & \textbf{Humans (1-5)} $\uparrow$ & & \textbf{\ours (\%)} $\downarrow$ & & \textbf{Humans (1-5)} $\uparrow$ & & \textbf{\ours (\%)} $\downarrow$ & & \textbf{Humans (\%)} $\downarrow$ \\
        \midrule
        HIVE~\cite{zhang2024hive} & & & \cellcolor{scale1}11.0 & & \cellcolor{scale1}2.0
        & & \cellcolor{scale1}40.5 & & \cellcolor{scale1}3.1
        & & \cellcolor{scale1}54.5 & & \cellcolor{scale1}40.5\\
        InstructPix2Pix~\cite{brooks2023instructpix2pix} & & & \cellcolor{scale2}15.5 & & \cellcolor{scale2}2.3
        & & \cellcolor{scale2}32.1 & & \cellcolor{scale2}3.3
        & & \cellcolor{scale2}44.0 & & \cellcolor{scale2}22.0\\
        MGIE~\cite{fu2024guiding} & & & \cellcolor{scale3}23.0 & & \cellcolor{scale3}2.7
        & & \cellcolor{scale4}21.6 & & \cellcolor{scale3}3.8
        & & \cellcolor{scale4}11.5 & & \cellcolor{scale4}10.0\\
        MagicBrush~\cite{zhang2023magicbrush} & & & \cellcolor{scale4}24.5 & & \cellcolor{scale4}2.9
        & & \cellcolor{scale3}23.1 & & \cellcolor{scale4}3.9
        & & \cellcolor{scale5}8.5 & & \cellcolor{scale5}5.5\\
        InstructDiffusion~\cite{geng2024instructdiffusion} & & & \cellcolor{scale5}30.0 & & \cellcolor{scale5}3.0
        & & \cellcolor{scale5}19.1 & & \cellcolor{scale5}4.0
        & & \cellcolor{scale3}13.5 & & \cellcolor{scale3}10.5\\
        \bottomrule
    \end{tabular}
    }
    \vspace{-0.15cm}
    \caption{Benchmark comparison of model rankings generated by \ours and those derived from the user study. The first two columns contrast average human ratings with scores obtained using \ours. The final column compares the percentage of unchanged images in the user study -- cases with maximal background preservation and minimal prompt adherence -- to the corresponding one identified by \ours.}
    \label{tab:benchmark}
    \vspace{-0.3cm}
\end{table*}

%% file: Tables/table_4_correlation_scores.tex
\begin{table}[t]
  \centering
  \setlength{\tabcolsep}{.4em}
  \resizebox{\linewidth}{!}{
  \begin{tabular}{lc ccc}
    \toprule
    & & $\rho$ & $\rho_s$ & $\tau$ \\
    \midrule
    \rowcolor{Gray}
    \multicolumn{5}{l}{\textit{Background Preservation}} \\
    CLIP-I & & 51.1 & 42.9 & 33.3 \\
    \hspace{0.3cm} w/ patch on random areas & & 34.2 & 25.6 & 19.4 \\
    \hspace{0.3cm} w/ patch on all detected differences & & 32.7 & 15.6 & 11.6 \\
    \rowcolor{OurColor}
    \hspace{0.3cm} w/ \textbf{\ours} (patch on coherent differences) & & \textbf{54.5} & \textbf{45.4} & \textbf{35.1} \\
    \midrule
    \rowcolor{Gray}
    \multicolumn{5}{l}{\textit{Prompt Adherence}} \\
    CLIP-T & & 21.5 & 23.1 & 17.1 \\
    \hspace{0.3cm} w/ patch on random areas & & 17.8 & 19.5 & 14.3 \\
    \hspace{0.3cm} w/ patch on all detected differences & & 13.3 & 14.6 & 10.7 \\
    \rowcolor{OurColor}
    \hspace{0.3cm} w/ \textbf{\ours} (patch on non-coherent differences) & & \textbf{24.6} & \textbf{26.6} & \textbf{19.6} \\
    \bottomrule
  \end{tabular}
  }
  \vspace{-0.2cm}
  \caption{Correlation analysis showing improved alignment with human judgment when integrating \ours in CLIP-I and CLIP-T.}
  \label{tab:correlation}
  \vspace{-0.4cm}
\end{table}

%% file: sections/5-conclusion.tex
\section{Conclusions}
In this work, we introduced \ours: a novel approach to evaluate instruction-guided image editing by detecting and assessing object-level differences between original and edited images. Our method leverages MLLMs to detect semantic differences between images and determine their coherence with the given editing instructions. This dual-step process allows for a structured and interpretable evaluation, addressing both the accuracy of the detected modifications and their semantic alignment with user intent.

%% file: sections/A-suppl.tex
\section{Difference Detection Training Dataset}
\tit{Stage 1: Additional Details}
To effectively train our difference detection model, we build our first stage on LVIS~\cite{gupta2019lvis}, which provides a broad range of annotated objects.
Given that LVIS encompasses 1,723 distinct categories, inconsistencies may arise during annotation, whereby annotators might label the same or similar objects under different categories or overlook certain objects. To address our objective of predicting missing objects in each pair, we filter them using the open-vocabulary detector OWL-ViT~\cite{minderer2022simple}. OWL-ViT is then tasked with verifying whether objects annotated in the first image (but not in the second) are genuinely absent in the second image and vice versa, thereby preventing erroneous ground-truth labels.

Moreover, considering the density of annotations in LVIS, we further refine our collection by eliminating annotations that are overly small, specifically, those with at least one dimension measuring less than 16 pixels. The final dataset consists of 118k images with a total of 795k, 725k, and 94k \texttt{ADD}, \texttt{REMOVE} and \texttt{EDIT} operations respectively.
 
\tit{Stage 2: Additional Details}
Building upon the previous stage, the second stage dataset is designed to refine our model for detecting differences in edited images.
The dataset is constructed using LVIS annotations, applying specific filtering criteria to ensure data quality. In detail, we sample a maximum of 4 objects per image. For each selected object, we consider its segmentation mask and filter out records that cover less than 3\% of the image area, while the maximum allowed overlap between selected object masks is 5\%.
As for the inpainting generation pipeline, the Kandinsky 2.2 inpaint model~\cite{razzhigaev2023kandinsky} is employed with 100 diffusion steps and a guidance scale of 4.

The final dataset comprises a total of 97k images for training and 19k images for testing. Within the training set, each of the three operations (\texttt{ADD}, \texttt{REMOVE}, and \texttt{EDIT}) is represented by approximately 33k instances to avoid unbalanced predictions. Additionally, 19k images remain unchanged. Similarly, in the test set, there are around 7k instances for each of the three operations, along with 5k images that remain unaltered.

\section{Training Details}
In each stage, the training of all the models is conducted within the same experimental setting. For instance, we apply QLoRA with rank 64, scaling factor 16, and a dropout rate of 0.1.
Both difference detection and coherence estimation fine-tuning employ 4-bit quantization and the paged AdamW 8-bit optimizer~\cite{loshchilovdecoupled}.

For image encoding, images are always center-cropped based on the smaller dimension to maintain a square aspect ratio. When employing Idefics3, images are internally resized to 1,456 pixels and encoded in a list of 364-pixel crops, generating a grid of 16 crops separately encoded. For Qwen2-VL, the image is encoded using a bidimensional positional embedding. Image tokens are then compressed by an MLP layer that encodes $2\times 2$ adjacent tokens into a single one. The input image resolution is kept at 1,456. For mPLUG-Owl3, the images are directly encoded through a SigLIP400m-384~\cite{zhai2023sigmoid}.

\tit{Difference Detection}
All stages of the difference detection training use 8 NVIDIA A100 64GB GPUs. Both training stages are conducted with a learning rate of $1 \times 10^{-4}$ and a total batch size of 8. In the first training stage, Idefics converges after 9k training steps, with evaluations performed every 1k steps. Differently, in the second training stage, convergence is reached after 30k training steps, with evaluations conducted every 5k steps.

\tit{Coherence Estimation} Training is performed on a single NVIDIA A100 64GB GPU with a batch size of one and a learning rate of $1 \times 10^{-5}$. In this case, Idefics reaches convergence at 550 training steps, with evaluations every 50 steps.

\begin{figure}[t]
    \centering
    \includegraphics[width=0.82\linewidth]{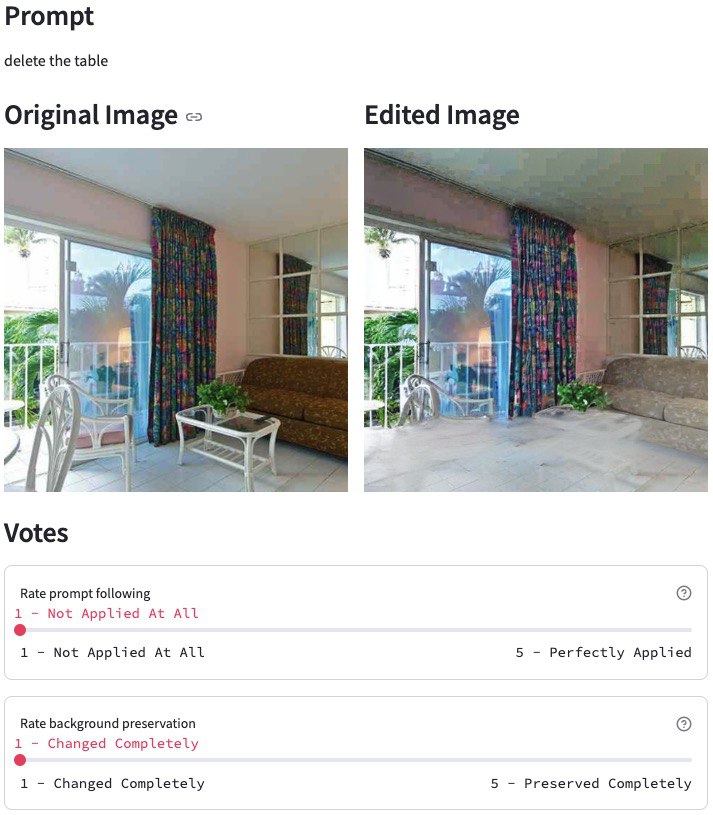}
    \vspace{-0.15cm}
    \caption{User study interface displaying the original and the edited image alongside the editing prompt.}
    \label{fig:user_study_platform}
    \vspace{-0.3cm}
\end{figure}

\section{User Study}
\label{sec:user_study}
To evaluate the performance of instruction-based image editing models, we conducted a user study in which participants assessed edited images based on prompt adherence and background preservation.

\tit{Data Generation} The editing models used for user study data generation are set according to their default configuration. In particular, for InstructDiffusion~\cite{geng2024instructdiffusion} the image guidance scale is set to 1.25 and the text guidance scale is set to 5, while for all the other models we use 1.5 and 7 respectively. For HIVE~\cite{zhang2024hive} and InstructDiffusion the number of inference steps is 100, while for all the other models we set the inference steps to 20.

Participants rated the degree to which the requested edit was applied on a scale from 1 to 5: \textit{not applied at all}, \textit{slightly applied}, \textit{moderately applied}, \textit{well applied}, and \textit{perfectly applied}. Similarly, they evaluated the preservation of background elements using another scale from 1 to 5: \textit{changed completely}, \textit{moderately altered}, \textit{mostly preserved}, \textit{nearly fully preserved}, and \textit{preserved completely}.
The study was conducted through an interactive interface that allowed participants to compare the original and the edited image while reading the prompt. Fig.~\ref{fig:user_study_platform} illustrates the interface used in the study.

\section{Additional Qualitative Results}

Fig.~\ref{fig:qualitatives_supp} illustrates a wide range of successful edits handled by \ours across diverse scenarios, featuring highly precise bounding boxes around the modified elements. In several examples, the system accurately detects color changes (\eg, altering the color of a bird or an umbrella) and clearly distinguishes added objects, such as a watermelon or a necktie. The pipeline also robustly identifies removed elements, whether it is a coffee table or a giraffe, and captures more subtle edit operations, for instance replacing a bird with a boy or a cat with a dog. These qualitative results highlight how \ours maintains spatial accuracy and class awareness when performing additions, removals, and edits, thereby demonstrating its adaptability to different editing requests.

Additionally, the coherence evaluator provides clear, step-by-step reasoning for each detected change, explaining why it is coherent or not. For example, as shown in Fig.~\ref{fig:qualitatives_reasoning}, in the ``change the color of the rose to blue'' prompt, the system points out that ``cyan'' is still a valid shade of blue and correctly labels the edit as coherent. Likewise, when asked to ``add a basketball to the top of the car'', it confirms not only that the basketball has been introduced but also that it is located precisely where the prompt requires -- on top of the car. These reasoned explanations highlight the transparency of our pipeline, helping users understand which aspects of the prompt were fulfilled and why certain edits might fall short of the requested modifications.

\section{Limitations}
\ours can occasionally exhibit inaccuracies in detected differences, as shown in Fig.~\ref{fig:failures}, that can end up affecting the coherence estimation. For instance, in the first example, changing the left hat to white actually matches the prompt. However, the pipeline labels the detected edit as ``fedora'', which confuses the coherence evaluator into classifying it as non-coherent. A similar problem arises in the second image, where the correct recoloring of the dog to brown is labeled as ``beige puppy'', causing confusion -- especially because the new color of the dog blends into the background.

Another failure can occur when the coherence evaluator is overly strict. In the ``replace dog with watermelon'' example, the editing model does replace the dog with a watermelon, yet the coherence evaluator rejects the result because it expects a total transformation. Similarly, in ``replace boy with girl,'' the difference detector notes that a person has been edited but fails to recognize the new person as female, and the coherence evaluator does not correct this oversight, causing it to label the modification as incoherent even though it actually meets the prompt requirements.

Further ambiguities in the prompt can lead to inaccurate predictions. Indeed, sometimes, the prompt does not specify all the elements needed to produce a univocal decision, especially when it is unclear whether the prompt specifies an addition or a substitution of elements in the scene.

\clearpage

\begin{table*}[t]
\centering
\footnotesize
\begin{tabular}{p{0.9\linewidth}}
\toprule
\textbf{Custom System Prompt:} \\
\texttt{You are a system that detects differences between two images.} \\
\\
\texttt{- Extract the elements that are changed in the second image with respect to the first one.} \\
\texttt{- Create a new entry for each distinct change.} \\
\texttt{- For each entry, use the following format:} \\
\texttt{"<CHANGE\_COMMAND>: <CHANGED\_ELEMENT>, (<BOUNDING\_BOX>)"} \\
\\
\texttt{CHANGE\_COMMAND:} \\
\texttt{- ADD: If a new element appears in the second image that was not present in the first.} \\
\texttt{- REMOVE: If an element from the first image is missing in the second.} \\
\texttt{- EDIT: If an element in the second image is different but in the same location as an element in the first image.} \\
\\
\texttt{CHANGED\_ELEMENT: Describe the element that has changed.} \\
\\
\texttt{BOUNDING\_BOX: Use normalized coordinates [x0, y0, x1, y1] for the changed element position in the second image, where (x0, y0) is the top-left corner, and (x1, y1) is the bottom-right corner. The coordinates should be scaled between 0 and 1, with 0 representing one edge of the image and 1 representing the opposite edge.} \\
\bottomrule
\end{tabular}
\vspace{-0.15cm}
\caption{Prompting template for the difference detection stage of the \ours pipeline.}
\label{tab:image_diff_prompt}
\end{table*}

\begin{table*}[t]
\centering
\footnotesize
\begin{tabular}{p{0.9\linewidth}}
\toprule
\textbf{Custom System Prompt:} \\
\texttt{You are evaluating if a specific change detected by an AI vision model matches the request in the original edit prompt.} \\
\\
\textbf{\#\# Task} \\
\texttt{Determine if the detected change, as described and bounded by the provided colored bbox, matches the request in the original edit prompt.} \\
\texttt{A match is valid only if the localized detected change is 100\% compatible with the requested prompt.} \\
\texttt{Any unwanted modification of the original image (even small) should avoid a match.} \\
\\
\textbf{\#\# Context} \\
- \texttt{The original image and the edited image are provided, in this order. The edited image is}\\
\texttt{the original with some changes applied. Focus only on the area specified by the bbox in the detected change.} \\
- \texttt{Another AI model has detected a change in the image, including its bbox.} \\
    - \texttt{ADD: An object is only added in the edited image (on the background).} \\
    - \texttt{EDIT: An object is substituted with another one in the edited image.} \\
    - \texttt{REMOVE: An object is removed in the edited image.} \\
- \texttt{Be strict: An EDIT means that an object has been removed and substituted with another one,}\\
\texttt{ensure nothing was removed unless explicitly stated in the prompt. If an object has been removed unexpectedly, then you should say NO.} \\
\\
\textbf{\#\# Example Response} \\
- \texttt{Reasoning: <REASONING>} \\
- \texttt{Decision: "YES" or "NO"} \\
\\
\textbf{User Prompt:} \\
\textbf{\#\# Instructions} \\
\texttt{1. The original edit prompt is: \{SUBSTITUTE\_PROMPT\}} \\
\texttt{2. The detected change to evaluate is: \{SUBSTITUTE\_CHANGE\}} \\
\texttt{3. Use only the text and the observations from the specified bbox area (colored) in both the} \\
\texttt{original and edited images to decide if the specific detected change aligns with the original edit prompt.} \\
\bottomrule
\end{tabular}
\caption{Prompting template for the coherence estimation stage of the \ours pipeline.}
\label{tab:eval_prompt_template}
\vspace{-0.15cm}
\end{table*}

%% file: sections/B-suppl.tex
\begin{figure*}[t]
    \centering
    \begin{tabular}{c}
    \includegraphics[width=\linewidth]{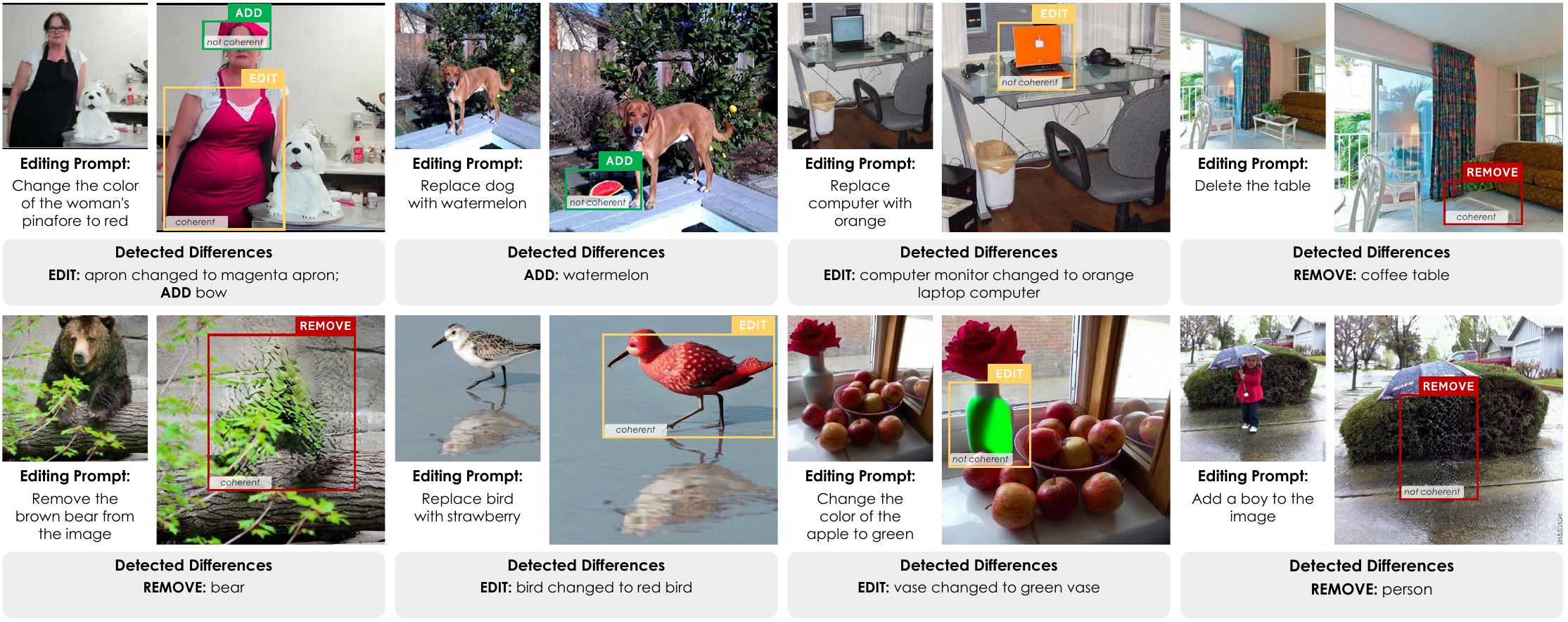}\\
    \addlinespace[0.08cm]
    \includegraphics[width=\linewidth]{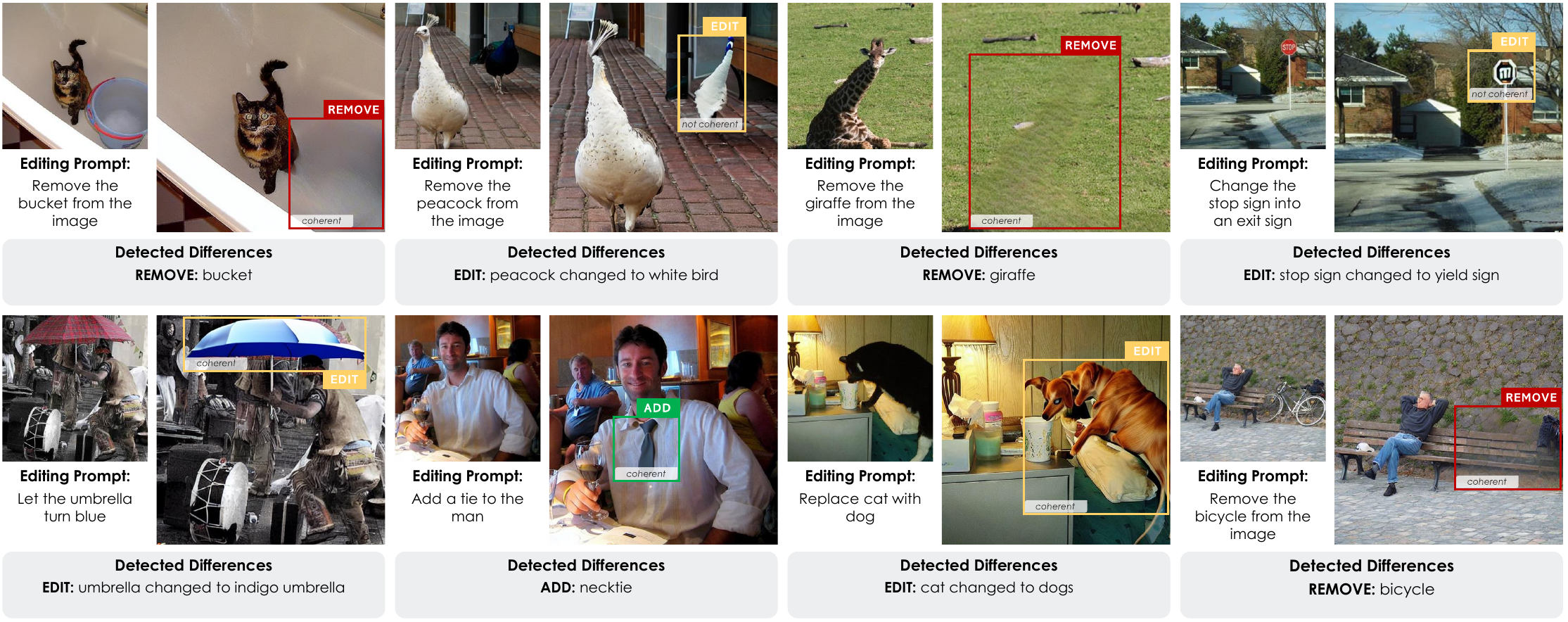}\\
    \addlinespace[0.08cm]
    \includegraphics[width=\linewidth]{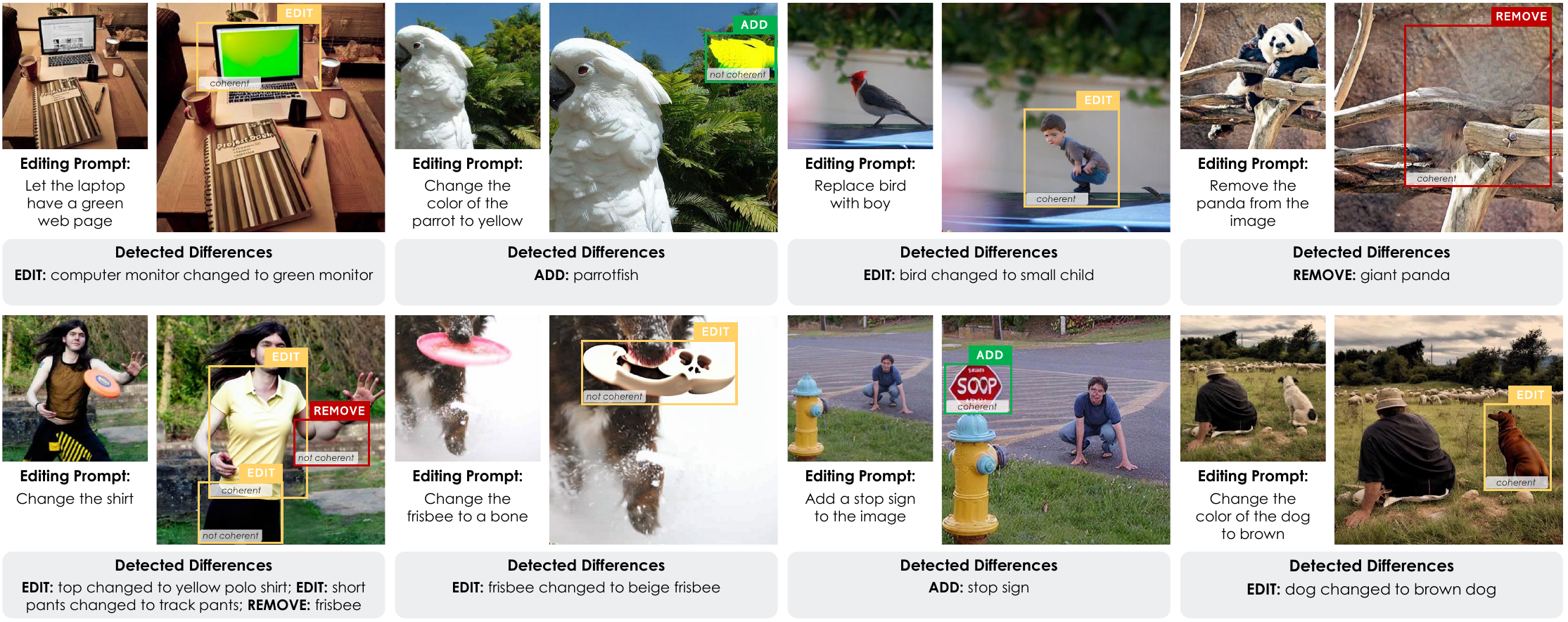}
    \end{tabular}
    \vspace{-0.15cm}
    \caption{Additional qualitative results. Each instruction-based edit shows the original image (left) and the edited version (right), alongside the given prompt.} 
    \vspace{-0.35cm}
    \label{fig:qualitatives_supp}
\end{figure*}

\begin{figure*}[t]
    \centering
    \includegraphics[width=\linewidth]{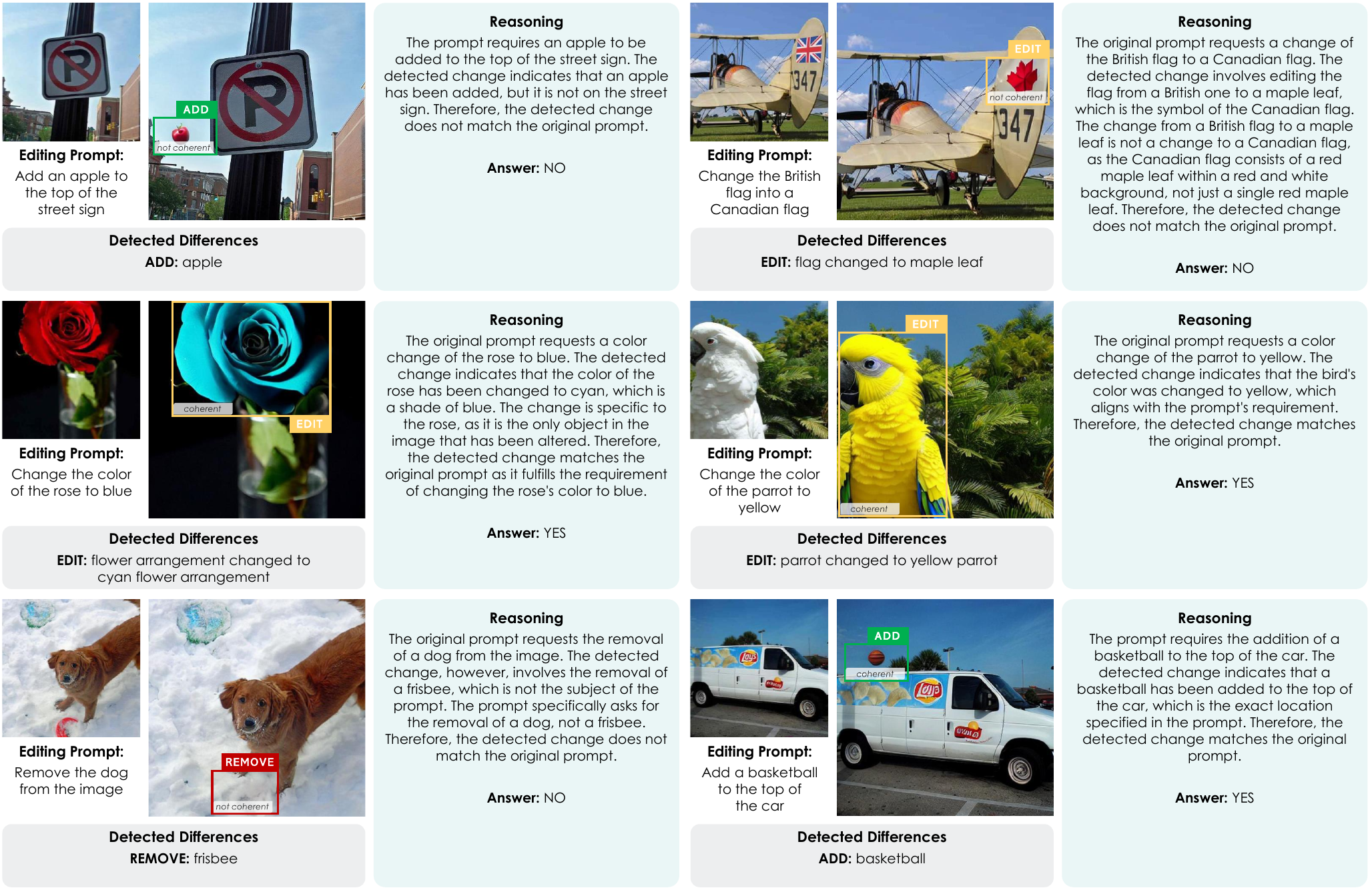}\\
    \vspace{-0.15cm}
    \caption{Examples illustrating the reasoning of the coherence evaluator that justifies its ‘YES’ or ‘NO’ decisions. Each box pairs a detected difference with an explanation of why the edit either fulfills or fails the user’s request, highlighting the ability of the pipeline to handle both spatial and semantic context.}
    \vspace{-0.35cm}
    \label{fig:qualitatives_reasoning}
\end{figure*}

\begin{figure*}[t]
    \centering
    \includegraphics[width=\linewidth]{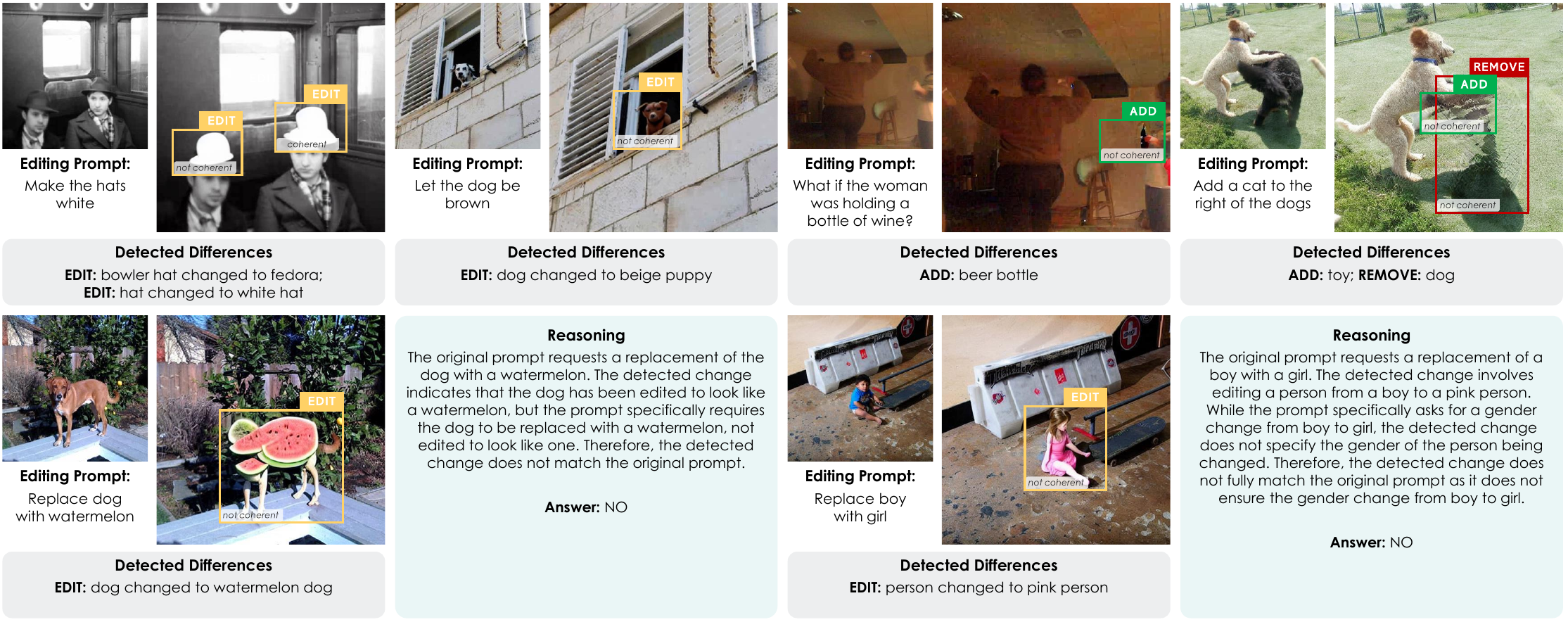}\\
    \vspace{-0.15cm}
    \caption{Failure cases where detection errors may impact coherence evaluation, potentially leading to misclassifications. Inaccurate identification of edits can introduce ambiguity for the coherence evaluator, while strict coherence criteria might occasionally reject valid changes, highlighting the interdependence of both stages.}
    \vspace{-0.35cm}
    \label{fig:failures}
\end{figure*}